# Facial Expression Recognition Based on Complexity Perception Classification Algorithm

**Tianyuan Chang , Guihua Wen\* , Yang Hu , JiaJiong Ma**

School of Computer Science and Engineering, South China University of Technology,Guangzhou, China
tianyuan_chang@163.com, crghwen@scut.edu.cn

## Abstract

Facial expression recognition (FER) has always been a challenging issue in computer vision. The different expressions of emotion and uncontrolled environmental factors lead to inconsistencies in the complexity of FER and variability of between expression categories, which is often overlooked in most facial expression recognition systems. In order to solve this problem effectively, we presented a simple and efficient CNN model to extract facial features, and proposed a complexity perception classification (CPC) algorithm for FER. The CPC algorithm divided the dataset into an easy classification sample subspace and a complex classification sample subspace by evaluating the complexity of facial features that are suitable for classification. The experimental results of our proposed algorithm on Fer2013 and CK+ datasets demonstrated the algorithm's effectiveness and superiority over other state-of-the-art approaches.

## 1 Introduction

Facial expression recognition (FER) has a wide range of research prospects in human computer interaction and affective computing, including polygraph detection, intelligent security, entertainment, Internet education, intelligent medical treatment, etc. As we all know facial expression is a major way of expressing human emotions. Hence, the main task in determining emotion is how to automatically, reliably, and efficiently recognize the information conveyed by facial expressions. In FER research, Ekman and Friensen first proposed the Facial Action Coding System (FACS) [Friensen and Ekman, 1983]. The six basic categories of expressions (surprise, sadness, disgust, anger, happiness, and fear) are defined in FACS, and are commonly used as the basic expression labels.

FER methods can be roughly divided into three types [Sun et al., 2017; Yan et al., 2017]: geometry-based methods, appearance-based methods, and hybrid methods. Geometry-based methods describe features mainly through geometric relationships such as key points on the face, face positions and shapes. Appearance-based methods extract the entire facial image as a whole. Lastly, hybrid methods combine the first two methods to extract local and global facial features of face images, respectively. Compared with appearance-based methods, the utility of geometry-based methods requires further improvement because it is difficult to accurately label the key points and effective positions of the face in practical applications.

Most FER works focus on feature extraction and classifier construction, which can be divided into static and dynamic classification methods [Corneanu et al., 2016]. Static classification methods are applicable for a single static image, and include SVM classifiers, Bayesian network classifiers, Random forests, and Softmax classifiers. Dynamic classification is applied to facial video sequences, taking into account features independently extracted from each frame over time as a basis; the main dynamic classification models are HMM [Le et al., 2011] and VSL-CRF [Walecki et al., 2015].

In recent years, different methods of traditional machine learning have been used in FER to extract appearance features of images, including Gabor filters, local binary patterns (LBP) [Savran et al., 2015], local Gabor binary patterns (LGBP) [Zhang et al., 2005], histograms of oriented gradients (HOG) [Dahmane et al., 2011], and scale-invariant feature transform (SIFT) [Berretti et al., 2011]. These traditional methods tend to be more effective in specific small sample sets, but are difficult to adjust to identify new face images, which brings challenges to FER in uncontrolled environments. This is because the extracted features often belong to low-level features, and it is hard to extract and find meaningful information that distinguishes the different categories from the data. However, a new recognition framework utilizing a convolutional neural network (CNN) [Ranzato et al., 2011; Liu et al., 2014; Kim et al., 2015] based on a deep learning network has achieved remarkable results in FER, and has been used for feature extraction and recognition. Multiple convolution and pooling layers in a CNN may extract higher and multi-level features of the entire face or local area, and has good classification performance of facial expression image features.

At present, there are still some deficiencies in FER. Many related work focus on the improvement of classification models and feature extraction methods, easily ignoring the relevance of several basic expression categories and the inconsistencies in the complexity of extracting features. As described by [Lopes et al., 2017], it is difficult to defini-tively partition each expression feature space. Some expressions, like happiness and surprise, belong to highly recognizable categories, which are easily distinguished by facial features. Meanwhile, other expressions, like fear and sadness, are very similar in some situations, making it difficult to distinguish them effectively. In addition, facial images are easily influenced by ethnicity, age, gender, hair, and other uncontrolled factors, resulting in different facial feature distributions and facial feature complexity for classification. Supposing that there are inconsistencies between the training samples and test samples, no matter how suitable the feature extraction of training samples in expression recognition, the prediction result of the test samples will not be accurate. An analogy of this is that we cannot ask an excellent student who has only grasped primary mathematics knowledge to be tested on knowledge of higher mathematics in a university. It is obvious that the complexity of the unequal knowledge needs to be distinguished for learning.

In order to effectively solve the above problems, in this paper we proposed a complexity perception classification (CPC) algorithm for FER referenced simple and effective principle. Our algorithm firstly divided the dataset into two parts: the difficult classification sub dataset, and the easy classification sub dataset. This was done by evaluating the complexity of facial features for expression recognition of each sample. The next step was to separately learn the two different feature distributions, that is, we trained two specific classifiers for these two sub datasets to obtain the complexity of different sample features. Instead of using a uniform classifier to predict the facial expression, our algorithm divided the test sample into corresponding classifiers to complete FER through the facial feature complexity discriminator.

Our main contributions can be summarized as follows.
1) According to the scale of the facial expression dataset, we proposed a simple and efficient CNN model based on ResNets for facial feature extraction, which effectively alleviated the problem of gradient disappearance, and enhanced information flow in the deep network.
2) A new heuristic classification algorithm named complexity perception classification (CPC) was proposed; it not only improved the recognition accuracy of higher recognizable facial expression categories, but also alleviated the problem of some misclassified expression categories.
3) We achieved state-of-the-art performance on static facial expression recognition benchmarks, including the Fer2013 dataset and CK+ dataset.

## 2 Feature Extraction

The classical improvement of CNN frameworks using ResNet [He et al., 2016] and DenseNet [Huang et al., 2016] not only alleviated the problem that deep networks are prone to gradient disappearance in backpropagation, but also remarkably improved the performance of image classification.

To this end and motivated by the above network architectures, we proposed a CNN framework for facial feature extraction as shown in Figure 1. Our CNN framework contained multiple convolutional layers, modified residual network blocks, and fully connected layers. In order to extract high level facial features, we selected the output of the second 1024 dimensional fully connected layer as the feature representation.

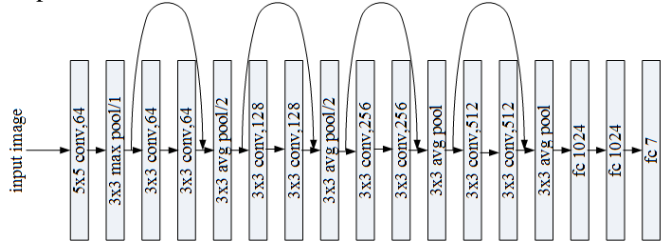

Figure 1: Pipeline of the CNN framework

In RestNets, the output of the feature mapping of the residual block consists of a non-linearly transformed composite function H(x) and an identity function x, which are combined as in Equation 1:

$$F(x) = H(x) + x \qquad (1)$$

This combination may hinder the flow of information through deep networks. In order to improve the flow of information between layers, we improved the combination mode of the residual block. Motivated by DenseNet, we no longer summated the two inputs, but concatenated the two feature mappings. The output function of the feature mapping is shown in Equation 2.

$$F(x) = [H(x), x] \qquad (2)$$

Figure 2 shows the structure of the traditional residual block and our modified residual block.

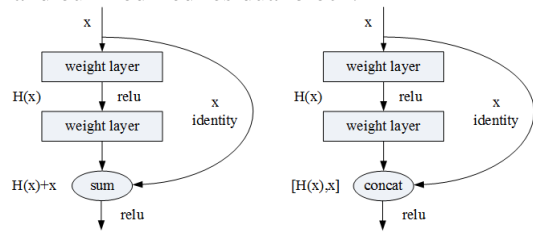

Figure 2: Left: Traditional residual block. Right: Modified residual block.

## 3 Complexity Perception Classification Algorithm for FER

Learning from the simple validity in Occam's razor principle, we proposed a complexity perception classification (CPC) algorithm to solve the complexity inconsistency problem of FER.

In this section, we firstly illustrated how to evaluate the complexity of training datasets for recognition. Then we employed the discriminant learing for different disjoint sub datasets and sample complexity discriminator to clearly determine the complexity of test sample for recognition. Figure 3 shows an overview and pipeline of the CPC system.

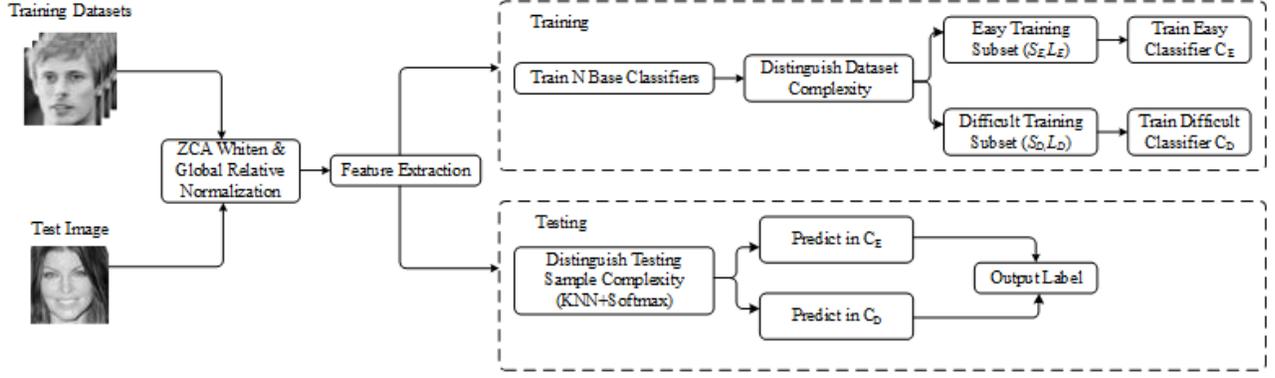

Figure 3: Pipeline of the complexity perception classification (CPC) system

### 3.1 Evaluating the Complexity of Features for Recognition

We adopted the feature extraction using the CNN as the input of the FER system for each feature value $f_i (i = 1,...,n)$, where $f_i$ is the i-th feature extracted from the original image. The feature extraction of training samples $X = [x_1, x_2, ..., x_n]$ were randomly divided into K folds, where $x_i \in R_d$, $i = 1, 2, ..., n$. $x_i$ is the i-th sample feature vectors and d is the feature dimension of each sample. In order to improve the generalization ability of the base classifier and to distinguish the easily classifiable samples from the difficultly classifiable samples, we first chose a fold of samples as the training set and the remaining K-1 folds of samples as the test set. This resulted in K base classifiers on different training datasets. We repeated the above process m times to obtain N(N = KM) trained base classifiers. Then, each training sample feature vector was predicted by n base classifiers, counting the correct number of prediction expression categories and computing the ease degree of classification $R(x_i)$ for the i-th sample:

$$R(x_i) = \frac{N(x_i)}{N} \quad (3)$$

where $N$ denotes the number of the base classifiers, and $N(x_i)$ is the correct classification number in N base classifiers of the i-th training sample.

We evaluated the complexity of sample classification features by the ease degree of classification $R(x_i)$. In addition, we set a parameter named ease threshold $\theta$ as the boundary to distinguish the easily classifiable samples from the difficultly classifiable samples.

$$x_i \in \begin{cases} S_E & \text{if } R(x_i) \geq \theta \\ S_D & \text{if } R(x_i) < \theta \end{cases} \quad (4)$$

As shown in Equation 4, when the ease degree of classification for the i-th training sample $R(x_i) \geq \theta$, we divided it into the easy classification sample subspace $(S_E, L_E)$. By contrast, we divided it into the difficult classification sample subspace $(S_D, L_D)$ when the ease degree of classification for the i-th training sample $R(x_i) < \theta$.

### 3.2 The Discriminant Learning

We divided the training datasets into two disjoint sub datasets by evaluating the feature complexity for recognition. In order to achieve discriminant learning in different sub datasets, we trained an easy sample classifier $C_E$ in the easy classification sample subspace $(S_E, L_E)$, and trained a difficult sample classifier $C_D$ in the difficult classification sample subspace $(S_D, L_D)$. In this way, we could learn different feature distributions in two different sample subspaces to provide more accurate recognition performance for the test sample. The easy sample classifier $C_E$ and the difficult sample classifier $C_D$ were as follows:

$$C_E = \xi(S_E, L_E), \quad (5)$$

$$C_D = \xi(S_D, L_D), \quad (6)$$

where $\xi$ is the specified classification method, such as Softmax, linear SVM, Random forest, etc. $S_E$ denotes the easy classification dataset, and $L_E$ denotes the expression labels of $S_E$. Similarly, $S_D$ denotes the difficult classification dataset, and $L_D$ denotes the expression labels of $S_D$.

### 3.3 Sample Complexity Discriminator

In order to be able to clearly determine whether the test sample belongs to the easy classification sample subspace or difficult classification sample subspace, we dynamically generated different sample complexity discriminators for different test samples. The sample complexity discriminator model is described as follows:

$$C_s = \psi(S_E \cup S_D, \{+,-\}), \quad (7)$$

where $\{+\}$ is the label of the easy classification dataset, $\{-\}$ is the label of the difficult classification dataset. The sample complexity discriminator model $C_s$ denotes a dynamic classification method of KNN-Softmax. We adopted the KNN algorithm to find k neighbors for the i-th test sample in the training set, and dynamically trained the sample complexity discriminator through the k nearest neighbor training set. The KNN-Softmax dynamic classification method could not only reflect the advantages of Softmax, but also improve the dis-crimination accuracy of the feature complexity for each test sample.

If the i-th test sample predicted {+} in the sample complexity discriminator, we used the easy sample classifier to recognize the facial expression. By contrast, if the label was {-}, we used the difficult sample classifier to recognize the facial expression.

Algorithm 1 elaborates the implementation method and the details of the CPC algorithm.

---

**Algorithm 1** Complexity perception classification (CPC)

---

**Input:** Feature vector $x_i$ from each sample, where $i \in 1,...,n$, classifier $\xi$, Complexity discriminator $\psi$, and ease threshold $\theta$.
**Output:** The predictive expression label $\omega$ for the testing sample.
**Training:**
1: Partition data into training, validation, and test sets.
2: **for** i = 1 to m **do**
3:   partition training data into K folds.
4:   **for** j = 1 to k **do**
5:     apply $\xi$ to train j-th training data, which gets a base classifier.
6:   **end for**
7: **end for**
8: Apply N base classifiers to compute the ease degree of classification $R(x_i)$ for each sample.
9: **for** each training sample $x_i$ **do**
10:   **if** $R(x_i) \geq \theta$
11:     Divide the sample $x_i$ into $S_E$.
12:   **else**
13:     Divide the sample $x_i$ into $S_D$.
14: **end for**
15: $C_E = \xi(S_E, L_E)$
16: $C_D = \xi(S_D, L_D)$
17: $C_I = \psi(S_E \cup S_D, \{+,-\})$ ]

**Testing:**
18: **for** each training sample **do**
19:   $y \leftarrow C_I(x)$
20:   **if** $y \in \{+\}$, $\omega \leftarrow C_E(x)$
21:   **if** $y \in \{-\}$, $\omega \leftarrow C_D(x)$
22: **end for**

---

## 4 Experiment

Our proposed algorithm was mainly applied to single-image static facial expression recognition. We evaluated the performance of the proposed complexity perception classification (CPC) algorithm by applying it to the Fer2013 [Goodfellow et al., 2013] and CK+ [Lucey et al., 2010] datasets.

### 4.1 Datasets

The Fer2013 dataset is a facial expression recognition challenge dataset that ICML 2013 launched on Kaggle. The dataset contains 28709 training images, 3589 validation images, and 3589 test images. The image size is $48 \times 48$, and facial expression is classified into seven different types (0=Angry, 1=Disgust, 2=Fear, 3=Happy, 4=Sad, 5=Surprise, 6=Neutral).

The CK+ dataset is an extension of the CK dataset. It contains 327 labeled facial videos, with each classified into one of the seven categories mentioned above, i.e., the same categories as those used in the Fer2013 dataset. We extracted four frames from each sequence in the CK+ dataset, which contains a total of 1308 facial expressions.

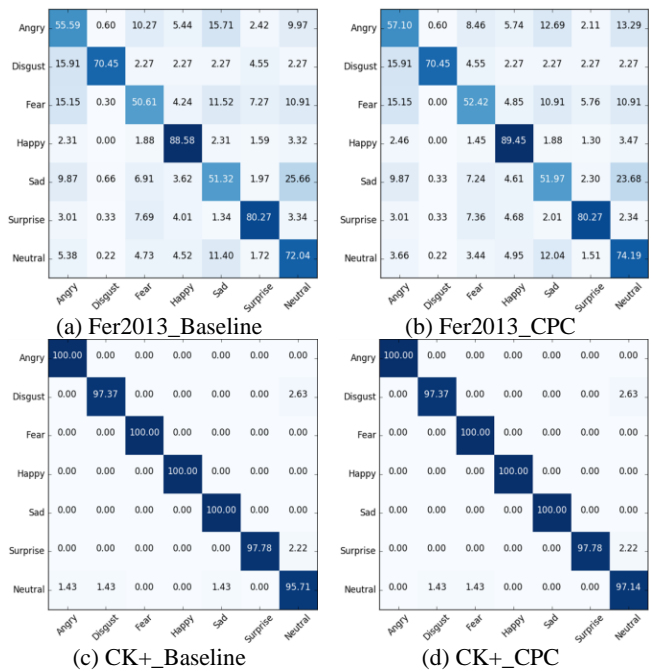

Figure 4: Confusion matrices for facial expressions on two datasets. The baseline is the result of feature extraction with the CNN predicted by Softmax. The CPC is the result of feature extraction with the CNN predicted by Softmax and adding the CPC algorithm.

### 4.2 Experimental Settings

Before the feature extraction phase in the CNN, we preprocessed the dataset with ZCA whitening and global relative normalization, which can effectively remove the redundant information of the input image and reduce the correlation between adjacent pixels in the image. In the pre-training phase of feature extraction with the CNN designed in Figure 1, we used the Fer2013 training dataset to train the networks. The initial network learning rate was set to 0.05, while the decay of the learning rate was per epoch. The mini batch size was 128, the momentum was set to 0.5, and the dropout was set to 0.2. The activation function used in the convolutional layers was the rectified linear unit (ReLU) activation function. The stochastic gradient descent (SGD) was used as the optimization algorithm. In addition, the last fully connected layer used Softmax as a multi-class activation function.

### 4.3 Baseline Classification vs. CPC

We firstly observed the impact of the CPC algorithm on the recognition accuracy of different expression categories. Figure 4 illustrates the resulting confusion matrices of testing accuracies (%) on the two datasets; we employed five-fold cross validation to ensure the stability of results. The effectiveness of the proposed algorithm was verified by compar-ing the effect of the complexity perception classification algorithm on recognition results after the feature extraction of the CNN.

On Fer2013 (Figure 4a), we can clearly see that the recognition rates of happiness were significantly higher than those of the other expressions belonging to the easily distinguishable category; meanwhile, the recognition rates of fear was the most difficult to distinguish. After adding the complexity perception classification algorithm to Fer2013 (Figure 4b), we found the following. (1) The recognition rate of happiness increased by 0.87% in the case of a high baseline recognition rate, the recognition rate of fear similarly increased by 1.81%, and the error rate related to mistaking fear for sadness decreased by 0.61%. These exciting recognition results demonstrated that the proposed algorithm not only improved the recognition accuracy of easily distinguishable categories, but also alleviated the easy misclassification of difficultly distinguishable categories. (2) The recognition rates of anger, sadness, and neutral also increased. The proposed algorithm improved the recognition rates for most categories. (3) None of the recognition rates of the expression classes decreased due to the addition of the proposed algorithm. Notice that our proposed algorithm did not raise the recognition accuracy of certain classes at the cost of sacrificing the accuracy of other classes, which is meaningful for practical application in facial expression recognition.

On CK+ (Figure 4c, 4d), the recognition accuracy of the natural class increased from 95.71% to 97.14%, while the other categories experienced no change in recognition accuracy upon adding the proposed algorithm. The results of the confusion matrix on the CK+ dataset also validate our above conclusion.

### 4.4 Performance of the CPC Algorithm

Table 1 show the recognition accuracy of different classifiers on the two datasets. We extracted facial features with the CNN, and compared the recognition effect of the CPC algorithm with the baseline recognition. In the experiment, we respectively trained two normal classifiers $C_n$ as the baseline recognition by employing the feature vectors of the training set on two datasets. The experimental results showed that the CPC algorithm had significant performance in facial expression recognition.

Compared to the baseline recognition rate on the Fer2013 test set, the average recognition rate of Softmax, linear SVM, and random forest classifiers were enhanced by 1.58%, 1.06%, and 0.59%, respectively, by adding the CPC algorithm. The average baseline recognition rate of the CK+ test set of the three different classifiers reached 97.76%. When we introduced the CPC algorithm, this average recognition increased to 98.27%.

### 4.5 Comparison of State-of-the-art Methods

We compared the proposed CNN feature extraction based on the CPC algorithm with existing related methods for FER. Table 2 and Table 3 respectively show the recognition accuracies of the different FER methods and our proposed algorithm on the CK+ and Fer2013 datasets. As can be seen in Table 2, our proposed algorithm achieved competitive results in the Fer2013 dataset. Table 3 shows that our proposed algorithm outperformed all other compared FER methods in the CK+ dataset.

Table 1: Comparison of recognition accuracies (%) of the baseline with CPC with different classifiers on the Fer2013 and CK+ datasets.

| Method | Fer2013 | | CK+ | |
|---|---|---|---|---|
| | Baseline | CPC | Baseline | CPC |
| CNN+Softmax | 69.77 | **71.35** | 98.38 | **98.78** |
| CNN+LinearSVM | 70.02 | **71.08** | 97.97 | **98.36** |
| CNN+RandomForest | 69.97 | **70.56** | 96.92 | **97.65** |

Table 2: Recognition rates (%) on the Fer2013 dataset for the state-of-the-art methods

| Method | Recognition rate (%) |
|---|---|
| Unsupervised [Goodfellow et al., 2013] | 69.26 |
| Maxim Milakov [Goodfellow et al., 2013] | 68.82 |
| [Mollahosseini et al., 2016] | 66.40 |
| [Tang et al., 2013] | 71.16 |
| [Liu et al., 2016] | 65.03 |
| DNNRL[Guo et al., 2016] | 70.60 |
| FC3072[Kim et al., 2016] | 70.58 |
| Proposed Approach | **71.35** |

Table 3: Recognition rates (%) on the CK+ dataset for the state-of-the-art methods

| Method | Recognition rate (%) |
|---|---|
| [Mollahosseini et al., 2016] | 93.20 |
| [Liu et al., 2014] | 94.19 |
| [Wu et al., 2015] | 98.54 |
| [Happy et al., 2015] | 94.09 |
| [Sun et al., 2016] | 94.87 |
| [Yan et al., 2017] | 96.60 |
| [Barman et al., 2017] | 98.30 |
| [Lopes et al., 2017] | 95.75 |
| [Sariyanidi et al., 2017] | 96.02 |
| Proposed Approach | **98.78** |

### 4.6 Parameter Analysis

We investigated the importance of the ease threshold $\theta$ parameter in the complexity perception classification algorithm. In order to determine the parameter value that produced the FER best performance on both datasets, we employed five-fold cross validation on the Fer2013 validation set and the CK+ dataset. Figure 5 and Figure 6 show the recognition accuracy of our algorithm versus different values of the ease threshold $\theta$ on two datasets, and compares the recognition accuracy with the baseline.

As can be seen in the above two figures, the recognition accuracy of our algorithm exceeded the baseline for most values of $\theta$, which illustrates the robustness of our algorithm. We also found that the random forest classifier was not stable enough due to its randomness. In addition, the relationship between recognition accuracy and different values of $\theta$ was similar for the linear SVM and Softmax classifiers.

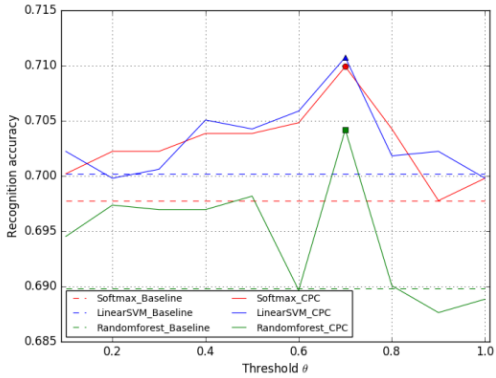

Figure 5: Recognition accuracy of the proposed algorithm versus different values of $\theta$ on the Fer2013 validation set.

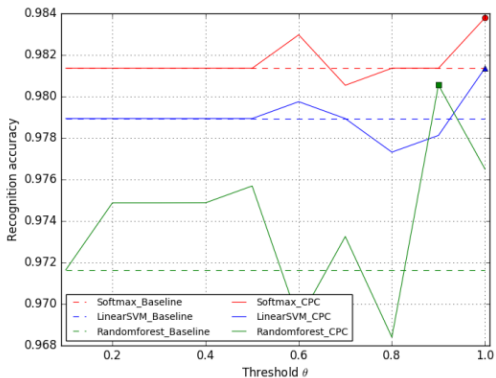

Figure 6: The recognition accuracy of the proposed algorithm versus different values of $\theta$ on the CK+ dataset.

## 5 Conclusions

In this paper, we designed a new and efficient CNN model for facial feature extraction, and proposed a complexity perception classification(CPC) algorithm for facial expression recognition. Our algorithm distinguished easily classifiable subspace from difficultly classifiable subspace to achieve discriminant learning of the facial feature distribution. Experimental results on the Fer2013 and CK+ datasets demonstrated that our algorithm outperformed state-of-the-art methods for facial expression recognition in terms of mean recognition accuracy.

## Acknowledgments

This work was supported by the China National Science Foundation (60973083, 61273363), the Science and Technology Planning Project of Guangdong Province (2014A010103009, 2015A020217002), and the Guangdong Science and Technology Planning Project (201504291154480).

## References

[Friensen et al., 1983] W. Friensen, P. Ekman, EMFACS-7: Emotional Facial Action Coding System, Technical Report, University of California, San Francisico, 1983.

[Sun et al., 2017] Wenyun Sun, Haitao Zhao, Zhong Jin. An Efficient Unconstrained Facial Expression Recognition Algorithm based on Stack Binarized Auto-encoders and Binarized Neural Networks. Neurocomputing, 2017.

[Yan et al., 2017] Haibin Yan. Collaborative discriminative multi-metric learning for facial expression recognition in video. Pattern Recognition, 2017.

[Corneanu et al., 2016] CA Corneanu, MO Simón, JF Cohn, et al. Survey on RGB, 3D, Thermal, and Multimodal Approaches for Facial Expression Recognition: History, Trends, and Affect-Related Applications. IEEE Transactions on Pattern Analysis & Machine Intelligence, 2016.

[Le et al., 2011] Vuong Le, H. Tang, and T.S. Huang. Expression recognition from 3D dynamic faces using robust spatio-temporal shape features. IEEE International Conference on Automatic Face & Gesture Recognition and Workshops IEEE, pages 414-421, 2011.

[Walecki et al., 2015] R. Walecki, O. Rudovic, V. Pavlovic, and M. Pantic, Variablestate latent conditional random fields for facial expression recognition and action unit detection, in Proc. IEEE Int. Conf. Autom. Face Gesture Recog., 2015.

[Savran et al., 2015] A. Savran, H. Cao, A. Nenkova, and R. Verma, Temporal Bayesian fusion for affect sensing: Combining video, audio, and lexical modali-ties, IEEE Trans. Cybern., pages 1927–1941, 2015.

[Dahmane et al., 2011] M Dahmane, J Meunier. Emotion recognition using dynamic grid-based HoG features. IEEE International Conference on Automatic Face & Gesture Recognition and Workshops IEEE, 2011.

[Berretti et al., 2011] S Berretti, BB Amor, M Daoudi. 3D facial expression recognition using SIFT descriptors of automatically detected keypoints. Visual Computer International Journal of Computer Graphics, 2011.

[Zhang et al., 2005] W Zhang, S Shan, X Chen. Local Gabor Binary Pattern Histogram Sequence (LGBPHS): A Novel Non-Statistical Model for Face Representation and Recognition." Tenth IEEE International Conference on Computer Vision IEEE, pages 786-791, 2005.

[Ranzato et al., 2011] M Ranzato, J Susskind, V Mnih G Hinton. On deep generative models with applications to recognition. Computer Vision and Pattern Recognition IEEE, pages 2857-2864, 2011.

[Liu et al., 2014] Mengyi Liu, R Wang, X Chen. Learning Expressionlets on Spatio-temporal Manifold for Dynamic Facial Expression Recognition. IEEE Conference on Computer Vision and Pattern Recognition IEEE Computer Society, pages 1749-1756, 2014.

[Kim et al., 2015] BK Kim, H Lee, J Roh. Hierarchical Committee of Deep CNNs with Exponentially-Weighted Decision Fusion for Static Facial Expression Recognition. ACM on International Conference on Multimodal Interaction ACM, pages 427-434, 2015:.

[Lopes et al., 2017] AT Lopes, ED Aguiar, AFD Souza. Facial expression recognition with Convolutional Neural Networks: Coping with few data and the training sample order. Pattern Recognition pages 610-628, 2017.

[He et al., 2016] K He, X Zhang, S Ren, J Sun. Deep Residual Learning for Image Recognition. Computer Vision and Pattern Recognition IEEE, pages 770-778, 2016.

[Huang et al., 2016] G Huang, Z Liu, LVD Maaten, KQ Weinberger. Densely Connected Convolutional Networks, 2016.

[Goodfellow et al., 2013] IJ Goodfellow, D Erhan, PL Carrier. Challenges in Representation Learning: A Report on Three Machine Learning Contests. International Conference on Neural Information Processing Springer, Berlin, Heidelberg, pages 117-124, 2013.

[Lucey et al., 2010] P Lucey, JF Cohn, T Kanade, J Saragihet. The Extended Cohn-Kanade Dataset (CK+): A complete dataset for action unit and emotion-specified expression. Computer Vision and Pattern Recognition Workshops IEEE, pages 94-101, 2010.

[Mollahosseini et al., 2016] Ali Mollahosseini, D. Chan, and MH Mahoor. Going deeper in facial expression recognition using deep neural networks. Applications of Computer Vision IEEE, pages 1-10, 2016.

[Tang et al., 2013] Y Tang. Deep Learning using Linear Support Vector Machines. Computer Science, 2013.

[Liu et al., 2016] Kuang Liu, M. Zhang, and Z. Pan. Facial Expression Recognition with CNN Ensemble. International Conference on Cyberworlds IEEE, pages 163-166, 2016.

[Guo et al., 2016] Yanan Guo, J Yu, H Xiong. Deep Neural Networks with Relativity Learning for facial expression recognition. IEEE International Conference on Multimedia & Expo Workshops IEEE Computer Society, pages 1-6, 2016.

[Kim et al., 2016] BoKyeong Kim, J Roh, and SY Dong. Hierarchical committee of deep convolutional neural networks for robust facial expression recognition. Journal on Multimodal User Interfaces, pages 1-17, 10.2, 2016.

[Wu et al., 2015] Chongliang Wu, S Wang, and Q Ji. Multi-instance Hidden Markov Model for facial expression recognition. IEEE International Conference and Workshops on Automatic Face and Gesture Recognition IEEE, pages 1-6, 2015.

[Happy et al., 2015] SL Happy, and A Routray. Automatic facial expression recognition using features of salient facial patches. IEEE Transactions on Affective Computing , pages 1-12, 6.1, 2015.

[Sun et al., 2016] Yaxin Sun, and Wen G. Cognitive facial expression recognition with constrained dimensionality reduction. Neurocomputing, 2016.

[Barman et al., 2017] Asit Barman, and P Dutta. Facial expression recognition using distance and shape signature features. Pattern Recognition Letters, 2017.

[Sariyanidi et al., 2017] Evangelos Sariyanidi, H Gunes, and A Cavallaro. Learning Bases of Activity for Facial Expression Recognition. IEEE Transactions on Image Processing. pages 1965-1978, 2017.